\documentclass{article}

    \PassOptionsToPackage{numbers, compress}{natbib}
\usepackage[preprint]{neurips_2026}
\usepackage[utf8]{inputenc} % allow utf-8 input
\usepackage[T1]{fontenc}    % use 8-bit T1 fonts
\usepackage{hyperref}       % hyperlinks
\usepackage{url}            % simple URL typesetting
\usepackage{booktabs}       % professional-quality tables
\usepackage{amsfonts}       % blackboard math symbols
\usepackage{nicefrac}       % compact symbols for 1/2, etc.
\usepackage{microtype}      % microtypography
\usepackage{xcolor}         % colors

\usepackage{titlesec}
\titlespacing*{\paragraph}{0pt}{0ex}{1em}
\usepackage{multirow}
\usepackage{graphicx}
\usepackage{amsmath}
\usepackage{xspace}
\usepackage{algorithm}      % floating algorithm environment
\usepackage{algpseudocode}  % modern algorithmicx pseudocode (recommended)

\newcommand{\ours}{TrajTok\xspace}

\author{%
  Zhen Xiong \\
  University of Southern California \\
  \texttt{z.xiong@usc.edu} \\
  \And
  Shang-Ling Hsu \\
  University of Southern California \\
  \texttt{hsushang@usc.edu} \\
  \AND
  Cyrus Shahabi \\
  University of Southern California \\
  \texttt{shahabi@usc.edu} \\
}

\title{\ours: Adaptive Spatial Tokenization for Trajectory Representation Learning}

\begin{document}

\maketitle

\begin{abstract}
Learning generalizable trajectory representations from raw GPS traces remains difficult because the data is continuous, noisy, irregularly sampled.
Even the choice of spatial tokenization is challenging: many existing approaches rely on fixed-size grids, but fine-resolution grids often produce cells with too few GPS visits to learn robust embeddings, while coarse-resolution grids suffer from internal fragmentation, merging heterogeneous local movement patterns into the same token.
We present \ours, a trajectory encoder with simple pretraining recipe that produce highly generalizable trajectory embeddings for diverse downstream applications. % quick background
First, we train a tokenizer that learns a mutli-resolution hexagonal cell partition from the spatial distribution of GPS points, mapping a continuous, noisy GPS sequence into a discrete series of cells as \textit{tokens}. Then, to capture both the geometric and kinematic features of trajectories, we introduce a factorized transformer encoder architecture with early per-modality self-attention blocks followed by cross-attention fusion layers. We further incorporate spatiotemporal rotary position embeddings (ST-RoPE) to encode where and when each token occurs along the trajectory. % arch intro
We pretrain TrajTok with a simple yet effecitve masked token modeling objective that encourages the model to recover both geometric structure and kinematics patterns from partial trajectory observations, leading to representations that can actually transfer effectively across diverse downstream tasks. % pretraining recipe
Empirically, our pretrained \textbf{\textit{frozen}} \ours\ encoder with lightweight task adapters reaches 0.435 HR@1 on trajectory similarity search, 0.773 macro-F1 on classification, 42.27\,s MAE on Estimated time of arrival (ETA), and 38.41\,s MAE on full travel-time regression on Porto dataset, outperforming multiple state-of-the-art task-specific methods. % empirical results
Notably, the \emph{same frozen} encoder handles both geometry-dominated tasks (e.g., similarity search) and kinematics-dominated tasks (e.g., travel-time estimation), indicating that the learned representation captures transferable structure rather than task-specific shortcuts. This suggests that our learned multi-resolution spatial tokenization, combined with simple masked-token pretraining, is a promising framework for general-purpose trajectory foundation models. % interpretation
\end{abstract}

\section{Introduction}

The analysis of human mobility has become a cornerstone of modern spatial computing, driving advancements in urban planning \cite{zheng2014urban}, epidemic control \cite{nature_mobility_covid}, and intelligent transportation systems \cite{didi_dispatching}. At the core of these applications lies the need to effectively model and understand GPS trajectories–sequences of geospatial coordinates that capture the continuous movement of entities over time. 

Despite the proliferation of massive spatial datasets, learning robust, generalized representations of GPS trajectories remains a formidable challenge. Raw trajectory data is inherently noisy, irregularly sampled, and strictly constrained by underlying road networks and map topologies. Existing foundation models for trajectory intelligence \cite{najjar2023towards, unitraj2025} and geographic representation learning paradigms, such as Space2Vec \cite{space2vec}, often rely on rigid, grid-based spatial discretization. These conventional methods struggle to preserve fine-grained spatio-temporal dynamics and fail to account for the complex structural constraints imposed by real-world environments.

To bridge this gap, we introduce \ours, a novel framework for learning geospatial representations of GPS trajectories via adaptive tokenization. Unlike traditional fixed-grid segmentation, our adaptive tokenization strategy partitions continuous trajectories into dynamical spatial tokens. Furthermore, to learn generalizable trajectory representations, \ours factorize geometric and kinematics features learning process at early stage using self-attention and then fuse them later via cross-attention mechanism.

To pretrain TrajTok, we adopt a simple factorized masked objective tailored to raw GPS token sequence. Given a partially masked trajectory, the geometric channel predicts the masked spatial cell IDs, while the kinematic channel reconstructs the corresponding motion targets, namely normalized speed and heading. Combined with density-adaptive tokenization, factorized geometric and kinematic representations, and cross-attention fusion with spatiotemporal positional encoding, this objective encourages the model to capture both route geometry and motion dynamics directly from noisy, irregularly sampled GPS traces. We evaluate the resulting pretrained encoder on four Porto benchmarks spanning both geometry-dominated and kinematics-dominated transfer: trajectory similarity search, call-type classification, prefix-based ETA prediction, and full-trajectory travel-time regression. Across all four tasks, a single frozen TrajTok encoder with lightweight task heads matches or outperforms strong prior task-specific methods, indicating that the learned representation captures transferable trajectory structure rather than task-specific shortcuts.

Our main contributions are summarized as follows:
\begin{itemize}
    \item We propose TrajTok, a trajectory representation framework that converts raw GPS trajectories into multi-resolution spatial tokens through a density-adaptive tokenization, allocating finer cells to dense regions while preserving a compact vocabulary.
    \item We introduce a factorized transformer encoder that separates geometric location signals from kinematic motion signals in early layers, then fuses them through cross-attention to form a unified trajectory representation.
    \item We develop a simple and effective pretraining recipe based on co-masked spatial token prediction and kinematic regression, together with run-aware span masking designed for raw token streams with repeated-cell runs.
    \item We show that the same frozen pretrained encoder transfers strongly across similarity retrieval, classification, estimated time of arrival (ETA) regression, demonstrating robust performance across both geometric and motion-sensitive downstream tasks.
\end{itemize}

\section{Related Work}

Some neural approaches to trajectory representation learning operate directly on raw coordinate sequences or point-level GPS features \cite{wang2018deepTTE, yang2025simformer}. Spatial discretization, however, remains widely used because it converts continuous, noisy GPS traces into reusable symbolic units amenable to embedding, reconstruction, masking, or contrast — instantiated in prior work as fixed grids, geohashes, road segments, or learned regions \cite{li2018deep, yao2019computing, deng2022efficient, chang2023contrastive, lin2023pretraining, najjar2023towards}. Yet because mobility data is strongly non-uniform, any fixed granularity faces an inherent tradeoff: fine cells preserve detail in dense areas but fragment sparse ones, while coarse cells offer stronger statistical support at the cost of merging heterogeneous patterns. This tension motivates adaptive tokenization, where spatial resolution is allocated according to the empirical density of trajectory observations.

A complementary line of work imposes stronger geographic structure through map matching and road-network modeling, representing trajectories as road-segment sequences and combining graph-based road encoders with transformer pretraining~\cite{jiang2023self, ma2024more, schestakov2024trajectory}. These approaches can be effective when accurate road graphs and matching pipelines are available, but rely on preprocessing assumptions that TrajTok deliberately avoids, operating instead on raw GPS trajectories after density-adaptive spatial tokenization.

Trajectory similarity learning has been a central evaluation protocol for the field, with methods such as t2vec \cite{li2018deep}, NeuTraj \cite{yao2019computing}, TrajCL \cite{chang2023contrastive}, and SIMformer \cite{yang2025simformer} defining strong baselines. At the same time, regression and classification tasks such as travel-time estimation and trip labeling provide complementary tests of whether a learned representation transfers beyond geometric similarity \cite{jiang2023self, schestakov2024trajectory}. 
% Our work follows this general-purpose representation-learning goal, but differs in its treatment of trajectory signals: \ours\ explicitly separates geometric location information from kinematic motion information in early layers, fuses the two through cross-attention, and pretrains on raw repeated-cell token streams with co-masked spatial prediction and kinematic regression.

\section{Methodology}
\label{sec:method}

\begin{figure}[tb]
    \centering
    \includegraphics[width=1.0\linewidth]{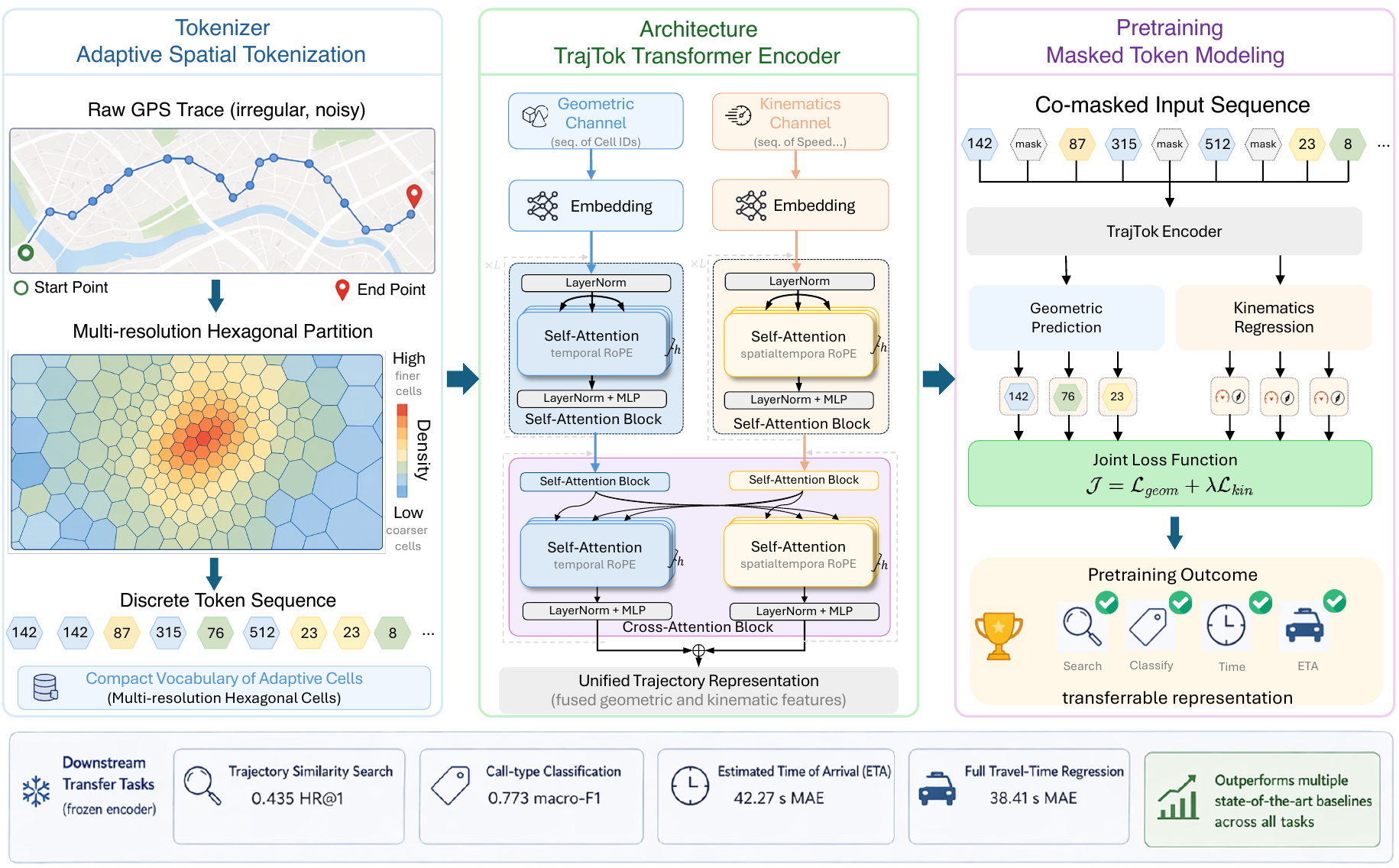}
    \caption{Overview of \textbf{TrajTok}. 
(\textbf{Left}) \emph{Adaptive spatial tokenization}: raw, irregular GPS trajectories are partitioned using a density-adaptive multi-resolution hexagonal grid, producing a compact vocabulary of spatial cells and a discrete token sequence. 
(\textbf{Middle}) \emph{Factorized transformer encoder}: tokens are represented through two streams—a geometric channel (cell IDs) and a kinematic channel (motion features such as speed and heading)—processed by separate self-attention blocks with spatiotemporal RoPE and fused via cross-attention to form a unified trajectory representation. 
(\textbf{Right}) \emph{Pretraining via co-masked token modeling}: masked positions are shared across channels, while geometric cell prediction and kinematic regression are optimized in parallel under a joint objective. 
The resulting pretrained encoder learns transferable representations that generalize across diverse downstream tasks.}
    \label{fig:overview}
\end{figure}

In this section, we describe the tokenization pipeline, the factorized encoder, and the masked pretraining objectives used by the current \ours\ implementation. See Figure~\ref{fig:overview} as the overview.

\subsection{Problem Formulation}

A raw GPS trajectory $\mathcal{T} = \{p_1, p_2, \ldots, p_n\}$ is a chronologically ordered sequence of points, where $p_i = (\phi_i, \lambda_i, t_i)$ contains latitude $\phi_i$, longitude $\lambda_i$, and timestamp $t_i$. Our goal is to learn an encoder $f_\theta(\mathcal{T})$ that produces contextual token states and a transferable trajectory representation suitable for both structure-oriented tasks such as similarity search and motion-sensitive tasks such as travel-time prediction.

\subsection{Adaptive Spatial Tokenization}
\label{sec:tokenization}

% \noindent
% \begin{minipage}[t]{0.48\textwidth}

\paragraph{Learning Density-adaptive vocabulary.}
We build the token vocabulary with the hierarchical H3 hexagonal grid~\citep{uberh3}. Starting from a base resolution $r_{\text{base}}$, we count training points inside each cell. Cells whose counts exceed a capacity threshold are split into their children, and the process repeats until the count falls below the threshold or the maximum resolution $r_{\text{max}}$ is reached. The resulting vocabulary contains cells at multiple resolutions and allocates finer spatial units to high-density regions. See Algorithm \ref{alg:adaptive_vocab} for more details.

\paragraph{Trajectory-to-tokens conversion.}
Each GPS point is mapped to the highest-resolution vocabulary cell that contains it. % TODO: a little bit more elbaboration?
A token therefore carries a discrete cell ID, the cell resolution, the original coordinates, and the timestamp. 
% We support both deduplicated and raw token streams. Deduplication removes consecutive repeated cells and shortens the sequence, but our strongest transfer backbone is trained on raw streams because they preserve dwell patterns and repeated-cell motion cues.

\subsection{Factorized Token Representation}
\label{sec:representation}

We represent each token along two factorized channels: a \emph{geometric} channel encoding where the token lies in space, and a \emph{kinematic} channel encoding how the agent is moving at that token. These two channels are processed by separate streams in the encoder (§\ref{sec:encoder}) and recombined through cross-attention fusion, allowing the model to specialize before unifying.

\paragraph{Geometric channel.}
The geometric embedding is a learned cell-identity vector,
\begin{equation}
    \mathbf{g}_j = \mathbf{e}_{c_j},
\end{equation}
where $\mathbf{e}_{c_j}$ is the embedding of the token's corresponding spatial cell.

\paragraph{Kinematic channel.}
The kinematic embedding captures per-token speed and heading. Rather than discretizing these quantities into bins, we represent them continuously and encode heading through its sine and cosine to respect its circular structure:
\begin{equation}
    \mathbf{x}^{\text{kin}}_j = \left[\frac{v_j}{v_{\max}}, \sin\theta_j, \cos\theta_j\right],
    \qquad
    \mathbf{k}_j = \mathrm{MLP}(\mathbf{x}^{\text{kin}}_j).
\end{equation}
Here $v_j$ is the haversine-distance-over-time speed between consecutive tokens. $\theta_j$ is the corresponding bearing; and the MLP is two-layer with GeGLU activation. 
% At masked kinematic positions, $\mathbf{k}_j$ is replaced with a learned mask embedding.

\subsection{Spatiotemporal RoPE}
\label{sec:encoder}

The encoder contains a geometric channel and a kinematic channel. Each channel first passes through independent embedding layer and then RoPE-equipped self-attention layers, after which paired fusion layers apply self-attention followed by cross-attention between channels.

\paragraph{Geometric positional encoding.} The geometric channel uses \textbf{S}patial-\textbf{T}emporal \textbf{Ro}tary \textbf{P}osition \textbf{E}mbeddings~\citep{su2024roformer} (ST-RoPE)  over latitude, longitude, and time. Coordinates and timestamps are normalized relative to the trajectory start.

\paragraph{Kinematic positional encoding.} The kinematic channel uses temporal-only RoPE, since speed and heading evolve primarily as functions of time. 
% Relative spatial context—the route geometry encoded by trajectory-relative coordinates—is injected into the kinematic channel through cross-attention from the geometric channel (described below), rather than by applying a second copy of spatial positional encoding to the kinematic channel.

\paragraph{Cross-attention fusion.} In a fusion block, each stream updates itself with commonly used self-attention and then attends to the other channel with cross-attention. For mathematical formulation of the self/cross-attention we used, see Appendix~\ref{app:attention} for details.

% \begin{align}
%     \tilde{\mathbf{G}}^{(\ell)} &= \mathrm{SelfAttn}_{\text{geo}}(\mathbf{G}^{(\ell)}), \\
%     \mathbf{G}^{(\ell+1)} &= \mathrm{CrossAttn}_{\text{geo}}(\tilde{\mathbf{G}}^{(\ell)}, \mathbf{K}^{(\ell)}),
% \end{align}
% with the kinematic stream updated symmetrically:
% \begin{align}
%     \tilde{\mathbf{K}}^{(\ell)} &= \mathrm{SelfAttn}_{\text{kin}}(\mathbf{K}^{(\ell)}), \\
%     \mathbf{K}^{(\ell+1)} &= \mathrm{CrossAttn}_{\text{kin}}(\tilde{\mathbf{K}}^{(\ell)}, \mathbf{G}^{(\ell)}),
% \end{align}

% Our best checkpoints use 16 total layers with 4 fusion layers, i.e., 12 independent layers per stream followed by 4 paired fusion blocks.

\subsection{Factorized Masked Pretraining}
\label{sec:pretraining}

\paragraph{Co-masked cell prediction.}
Given a mask set $\mathcal{M}$, the geometric stream predicts the original cell IDs at masked positions:
\begin{equation}
    \mathcal{L}_{\text{geom}} =
    - \frac{1}{|\mathcal{M}|}
    \sum_{j \in \mathcal{M}} \log p_\theta(c_j \mid \mathbf{G}).
\end{equation}

\paragraph{Co-masked kinematic regression.}
At the same masked positions, the kinematic stream reconstructs normalized speed and heading targets:
% \begin{equation}
%     \mathbf{y}_j =
%     \left[\frac{v_j}{v_{\max}}, \sin(\theta_j), \cos(\theta_j)\right].
% \end{equation}
% The regression loss is
\begin{equation}
    \mathcal{L}_{\text{kin}}
    =
    \beta_{\text{speed}} \, \mathrm{MSE}(\hat{v}_j, v_j / v_{\max})
    +
    \frac{1}{2}\beta_{\text{heading}} \mathrm{MSE}(\widehat{\sin \theta}_j, \sin \theta_j)
    +
    \frac{1}{2}\beta_{\text{heading}} \mathrm{MSE}(\widehat{\cos \theta}_j, \cos \theta_j)
\end{equation}

\paragraph{Joint objective.}
The default joint pretraining objective we use is the linear combination of previous two loss functions: 
\begin{equation}
    \mathcal{J} = \mathcal{L}_{\text{geom}} + \lambda \mathcal{L}_{\text{kin}}.
\end{equation}

We use \textit{run-aware} span masking that rejects candidate spans lying entirely inside a single repeated-cell run; see Appendix~\ref{app:abl-mask} for details and a comparison to naive masking.

\section{Experiments}
\label{sec:experiments}

We evaluate a single frozen \ours\ encoder across four trajectory-related tasks spanning two regimes: geometry-dominated (§\ref{sec:sim}: trajectory similarity search , §\ref{sec:cls}: call-type classification) and kinematics-dominated (§\ref{sec:eta}: ETA from prefixes, §\ref{sec:tte}: full-trajectory travel-time regression). With only lightweight task adapters, our frozen pretrained encoder matches or outperforms task-specific baselines across all four benchmarks (Tables~\ref{tab:sim}--\ref{tab:tte}). Controlled ablations (§\ref{sec:ablations}) further isolate three drivers of this transfer: density-adaptive tokenization, cross-attention fusion between factorized channels, and pretraining recipe.

\subsection{Experimental Setup}
\label{sec:setup}

For dataset statistics, tokenization configuration, spatial vocabulary and model hyperparameters, see Appendix~\ref{app:setup} for detailed description.

\begin{figure}[!htbp]
    \centering
    \includegraphics[width=1\linewidth]{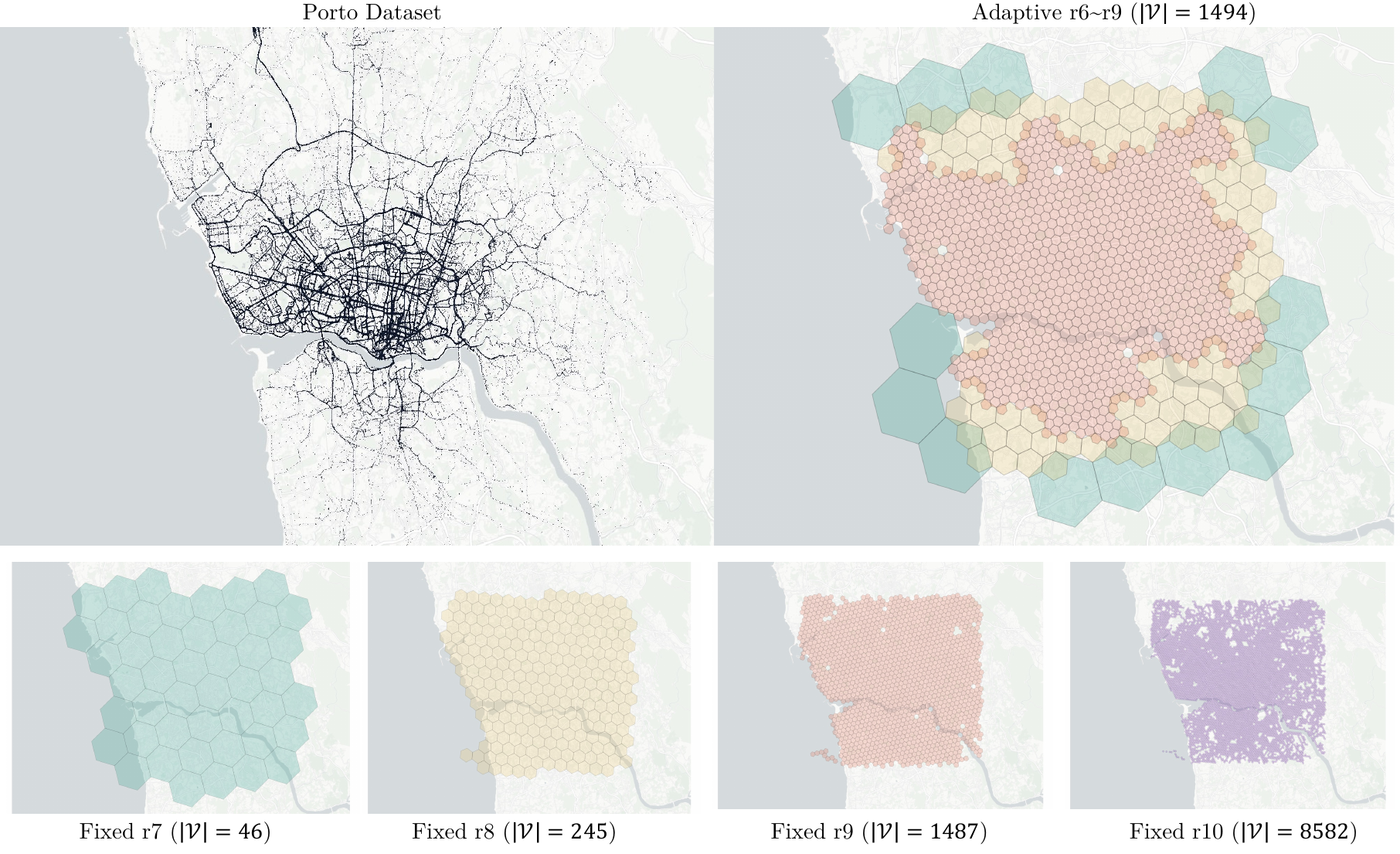}
    % \caption{Comparison between fixed-resolution and adaptive spatial tokenization on the Porto map. The top-left panel shows the raw Porto dataset, while the top-right panel illustrates the proposed adaptive partition (levels $r6$--$r9$, $|\mathcal{V}|=1494$), where cell density varies according to spatial coverage and data distribution. The bottom row presents fixed-resolution hexagonal grids at different levels ($r7$--$r10$), with vocabulary sizes ranging from $46$ to $8582$. Compared to uniform discretization, the adaptive scheme achieves a more balanced trade-off between spatial resolution and vocabulary size by allocating finer granularity in dense regions and coarser cells in sparse areas.}
    \caption{Fixed vs. adaptive spatial tokenization on Porto. Top: raw data (left) and adaptive partition (right; $r6$–$r9$, $|\mathcal{V}|=1494$), where cell density follows data distribution. Bottom: fixed hex grids using different resolutions. Adaptive tokenization better balances resolution and vocabulary by refining dense regions and coarsening sparse areas.}
    \label{fig:tokenization}
\end{figure}

% \paragraph{Frozen transfer protocol.} Throughout §\ref{sec:sim}--§\ref{sec:eta}, the pretrained encoder is frozen and only lightweight task heads are trained. Consistent with prior work \cite{jiang2023self,schestakov2024trajectory}, task heads may incorporate standard task-specific context features (e.g., departure metadata for classification, known destination for ETA); what transfers is the encoder representation of the observed trajectory.

\subsection{Trajectory Similarity Search}
\label{sec:sim}

We construct a DTW\footnote{Dynamic Time Warping (DTW) is an algorithm for measuring similarity between two temporal sequences}-labeled retrieval bank from the held-out test split: 1{,}000 queries against a fixed 10{,}000-trajectory corpus, with ground-truth relevance defined by DTW distance on raw GPS sequences. Top-$k$ Hit Ratio (HR@$k$) and Mean Reciprocal Rank (MRR) evaluate whether the embedding-based ranking retrieves the DTW-nearest trajectory, treating the single exact-DTW nearest neighbor as the primary positive;  top-$k$ Partial Hit Ratio (R$m$@$k$) and Normalized Discounted Cumulative Gain (NDCG) further evaluate how well the retrieved ranking preserves the top exact-similarity neighbors~\cite{long2026region}. Ranking is by cosine similarity of L2-normalized embeddings. Ranking is by cosine similarity of L2-normalized embeddings. Our
retrieval head is finetuned with mixed InfoNCE and rank distillation; see Appendix~\ref{app:adapters} for the full configuration.

\paragraph{Results.} Table~\ref{tab:sim} reports both a zero-shot variant (mean-pooled frozen encoder output) and the attentive retrieval head. With the retrieval head, \ours\ attains the best HR@1 (0.435), R5@20 (0.983), and MRR (0.588), improving over the strongest prior baseline by 0.127, 0.176, and 0.140 respectively. On NDCG@50, \ours\ ranks second to NeuTraj (0.658 vs.\ 0.672); we note that our retrieval head's training criterion (Appendix~\ref{app:adapters}) emphasizes top-of-list correctness over the broader top-50 ordering that NDCG@50 rewards. Additionally, our zero-shot variant alone already surpasses all prior methods on HR@1 and R5@20, indicating that the pretrained representation captures DTW-aligned geometric structure without any retrieval-specific finetuning.

\begin{table}[!tb]
\centering
\caption{Trajectory similarity search on Porto. Both \ours\ rows share a common retrieval bank and DTW ground truth. Best results in \textbf{bold}; Second best are \underline{underlined.}}
\label{tab:sim}
\small
\begin{tabular}{lcccc}
\toprule
Method & HR@1 $\uparrow$ & R5@20 $\uparrow$ & MRR $\uparrow$ & NDCG@50 $\uparrow$ \\
\midrule
t2vec \cite{li2018deep} & 0.239 & 0.624 & 0.373 & 0.490 \\
CL-TSim \cite{deng2022efficient} & 0.116 & 0.362 & 0.204 & 0.266 \\
NeuTraj \cite{yao2019computing} & 0.308 & 0.807 & 0.448 & \textbf{0.672} \\
TrajCL \cite{chang2023contrastive} & 0.217 & 0.603 & 0.343 & 0.556 \\
TrajGAT \cite{yao2022trajgat} & 0.296 & 0.683 & 0.437 & 0.450 \\
KGTS \cite{chen2024kgts} & 0.253 & 0.444 & 0.369 & 0.411 \\
SIMformer \cite{yang2025simformer} & 0.300 & 0.755 & 0.446 & 0.651 \\
Space2Vec \cite{mai2020multi} & 0.260 & 0.879 & 0.380 & 0.291 \\
SatClip \cite{klemmer2025satclip} & 0.177 & 0.907 & 0.300 & 0.406 \\
\midrule
\ours\ (zero-shot)       & \underline{0.351} & \underline{0.964} & \underline{0.503} & 0.528 \\
\ours\ (retrieval head)  & \textbf{0.435} & \textbf{0.983} & \textbf{0.588} & \underline{0.661} \\
\bottomrule
\end{tabular}
\end{table}

\subsection{Trajectory Classification}
\label{sec:cls}

We predict Porto \emph{call type} (different labels indicating how the
taxi trip was requested). The classifier freezes the encoder and trains
a head that concatenates attentive-pooled trajectory features with a
compact departure-context information; see Appendix~\ref{app:adapters} for
adapter and optimization details.

\paragraph{Results.}
Table~\ref{tab:cls_call} shows that \ours\ attains the best
macro-F1 (0.773) and the second-best micro-F1 (0.811), within
0.002 of START (0.813 / 0.772). We do not claim a decisive
numerical advantage on this benchmark in isolation; the relevant
comparison is qualitative. START relies on map-matched road
segments and a dedicated road-network encoder, and JGRM (0.805 /
0.768) jointly models GPS points and routes — both incorporate
stronger geographic inductive biases than \ours, which operates
only on raw GPS streams. Matching these specialized pipelines
with a single \emph{frozen} encoder and a lightweight head
supports our claim that the learned representation captures
transferable trajectory structure rather than task-specific
shortcuts. Because call type is a geometry-dominated label, this
result primarily exercises the effectiveness of our geometric channel.
%§\ref{sec:eta}–§\ref{sec:tte} provide the kinematic complement.

\begin{table}[h]
\centering
\caption{Porto call-type classification. Best results in \textbf{bold}; Second best are \underline{underlined.}}
\label{tab:cls_call}
\small
\begin{tabular}{lcc}
\toprule
Method & Micro-F1 $\uparrow$ & Macro-F1 $\uparrow$ \\
\midrule
Traj2vec~\cite{yao2018learning} & 0.515 & 0.323 \\
TrajCL~\cite{chang2023contrastive}   & 0.801 & 0.761 \\
PIM~\cite{yang2021unsupervised}      & 0.728 & 0.613 \\
START~\cite{jiang2023self}    & \textbf{0.813} & \underline{0.772} \\
JGRM~\cite{ma2024more}     & 0.805 & 0.768 \\
\ours\ (ours) & \underline{0.811} & \textbf{0.773} \\
\bottomrule
\end{tabular}
\end{table}

% ETA application
\subsection{Estimated Time of Arrival (ETA) from Prefixes}
\label{sec:eta}

ETA is our primary kinematic transfer test. Given an observed
\emph{prefix} of a trip and a known destination, the model predicts the
remaining travel time. The frozen encoder consumes the prefix
trajectory, and a destination-conditioned adapter predicts the remaining
travel time under a randomized-prefix training regime; full adapter
configuration is in Appendix~\ref{app:adapters}.

% To contextualize the difficulty of Porto-ETA, we compare with published neural travel-time systems evaluated on Porto, including en-route methods that estimate the remaining travel time after observing a partial trip \cite{wang2018deepTTE,fu2020compacteta,fang2020constgat,fang2021ssml,fan2022metaer}.

\paragraph{Results.}
Table~\ref{tab:eta_transfer} shows that \ours\ achieves
\textbf{42.27\,s} MAE on randomized-prefix ETA, substantially below the
reported Porto MAE range of prior neural travel-time systems. This result
is notable because \ours\ uses a single frozen representation shared across
downstream tasks, with no task-specific trajectory encoder and no manual
feature engineering. 
% The strong $R^2$ of \textbf{0.865} further indicates that the learned representation preserves destination-conditioned kinematic information from partial trajectories.

\begin{table}[t]
\centering
\caption{Estimated Time of Arrival on Porto. \textbf{Bold} marks the best reported value.}
\label{tab:eta_transfer}
\small
\begin{tabular}{lcccc}
\toprule
Method & MAPE (\%) $\downarrow$ & MAE (s) $\downarrow$ & RMSE (s) $\downarrow$ \\
\midrule
DeepTTE~\cite{wang2018deepTTE}        & 19.65 & 119.52 & 183.59 \\
CompactETA~\cite{fu2020compacteta}    & 19.37 & 115.82 & 174.25 \\
ConSTGAT~\cite{fang2020constgat}      & 18.68 & 109.79 & 168.39 \\
TransferTTE~\cite{fan2022metaer}      & 18.61 & 109.56 & 168.17 \\
MAML~\cite{finn2017maml}              & 17.80 & 102.02 & 158.76 \\
SSML~\cite{fang2021ssml}              & 17.68 &  98.63 & 150.96 \\
MetaER-TTE~\cite{fan2022metaer}       & 17.37 &  93.75 & 145.39 \\
\ours\ (ours)                         & \textbf{12.42} & \textbf{42.27} & \textbf{72.05} \\
\bottomrule
\end{tabular}
\end{table}

% TTE application
\subsection{Full-Trajectory Travel-Time Regression}
\label{sec:tte}

As a complementary kinematic benchmark, we predict total travel time from the entire observed trip. We acknowledge that total travel time admits a strong correlation with observed sequence length, which makes this task substantially easier than prefix-based ETA.

\paragraph{Results.} Table~\ref{tab:tte} eports total travel-time MAE from the full observed trajectory. This is a strict subcase of ETA at prefix=100\% and admits a strong length correlation, so it is a weaker test than §4.4. Nonetheless, TrajTok still attains 38.4\,s MAE, a substantial gap to prior work.

\begin{table}[h]
\centering
\caption{Full-trajectory travel-time regression on Porto. See §\ref{sec:tte} for protocol caveat.}
\label{tab:tte}
\small
\begin{tabular}{lccc}
\toprule
Method & MAE (s) $\downarrow$ & MAPE (\%) $\downarrow$ & RMSE (s) $\downarrow$ \\
\midrule
t2vec \cite{li2018deep}                 & 115.27 & 17.68 & 172.83 \\
CL-TSim \cite{deng2022efficient}        & 162.62 & 25.30 & 241.67 \\
TrajCL \cite{chang2023contrastive}      &  97.42 & 14.38 & 154.32 \\
CSTTE \cite{lin2023pretraining}         & 113.29 & 17.04 & 175.52 \\
START \cite{jiang2023self}              & 131.40 & 19.94 & 196.71 \\
TIGR \cite{schestakov2024trajectory}    &  86.86 & 12.76 & 139.80 \\
\ours\ (ours)                           & \textbf{38.41} & \textbf{5.71} & \textbf{65.90} \\
\bottomrule
\end{tabular}
\end{table}

\subsection{Ablation Studies}

\label{sec:ablations}

We ablate \ours\ along three axes corresponding to our three core design choices: density-adaptive tokenization (§\ref{sec:abl_tok}), cross-attention fusion between factorized channels (§\ref{sec:abl_fusion}), and the pretraining recipe (§\ref{sec:abl_pretrain}). Each ablation isolates a single design decision while holding all other components fixed. Additional controls, including masking-strategy variants and deduplication compression, are discussed and reported in Appendix~\ref{app:abl}.

\subsubsection{Density-Adaptive Tokenization}
\label{sec:abl_tok}

We first study whether adaptive tokenization is necessary or if a well-chosen fixed resolution suffices. Table~\ref{tab:abl_tokenizer} compares our density-adaptive vocabulary against fixed resolutions from 7 to 10 (see Figure~\ref{fig:tokenization} for visualization), with all other components held constant.

\begin{table}[h]
\centering
\caption{Tokenizer ablation: density-adaptive vs.\ fixed-resolution H3. All configurations use the same factorized encoder, factorized masked objective, and 12k training steps. Retrieval metrics come from zero-shot DTW similarity search on a 1k query $\times$ 10k corpus Porto bank (frozen encoder, no finetuning). Best results in \textbf{bold}.}
\label{tab:abl_tokenizer}
\small
\begin{tabular}{lcccccc}
\toprule
Tokenizer & Vocab size & Val loss $\downarrow$ & HR@1 $\uparrow$ & MRR $\uparrow$ & NDCG@5 $\uparrow$ & Spearman $\uparrow$ \\
\midrule
r7 (\textit{fixed})  & 46      & 0.143\textsuperscript{\dag} & 0.197 & 0.314 & 0.296 & 0.263 \\
r8 (\textit{fixed})  & 245     & 0.330 & 0.275 & 0.421 & 0.443 & 0.470 \\
r9 (\textit{fixed})  & 1{,}487 & 0.333 & \textbf{0.293} & 0.414 & 0.428 & 0.547 \\
r10 (\textit{fixed}) & 8{,}582 & 0.752 & 0.259 & 0.393 & 0.406 & 0.494 \\
\midrule
adaptive (\textit{ours}) & 1{,}494 & \textbf{0.297} & 0.286 & \textbf{0.424} & \textbf{0.445} & \textbf{0.548} \\
\bottomrule
\end{tabular}
\begin{flushleft}
\footnotesize\textsuperscript{\dag}r7 defines a trivial 46-class prediction task, so its low val loss is not comparable across resolutions; the retrieval columns is a more fair indicator.
\end{flushleft}
\end{table}

The density-adaptive vocabulary achieves the strongest overall retrieval profile, with the best comparable validation loss, MRR, NDCG@5, and Spearman correlation. We note that fixed r9 is negligibly higher on HR@1 (0.293 vs.\ 0.286) --- a difference of seven queries in the 1k-query bank. Since HR@1 depends on the single exact-DTW nearest neighbor, this suggests that uniform r9 can occasionally better recover the exact top match in dense regions. However, the adaptive tokenizer ranks the broader neighborhood more consistently, as reflected by its stronger MRR, NDCG@5, and Spearman scores. Coarser resolutions (r7, r8) lack spatial precision, while finer resolution (r10) fragments the vocabulary and weakens statistical support per cell. The adaptive scheme resolves this tradeoff by allocating fine cells where data density warrants them, a capacity no single fixed resolution can match.

\subsubsection{Cross-Attention Fusion Between Factorized Channels}
\label{sec:abl_fusion}

Our architecture factorizes trajectory representation into a geometric channel and a kinematic channel, coupled by cross-attention fusion layers. We test whether each component of this design is necessary by evaluating four architectural variants across all four downstream tasks.

\begin{table}[h]
\centering
\caption{Architectural ablation across all four downstream tasks. Similarity reports HR@1; classification reports macro-F1; ETA and TTE report MAE in seconds (lower is better). Best results in \textbf{bold} and second best results are \underline{underlined}.}
\label{tab:abl_fusion}
\small
\begin{tabular}{lcccc}
\toprule
Variant & Similarity $\uparrow$ & Classification $\uparrow$ & ETA $\downarrow$ & TTE $\downarrow$ \\
\midrule
Geometric (only)      & \underline{0.286} & \underline{0.773} & 44.44 & 87.52 \\
Kinematic (only)      & 0.028 & 0.632 & \underline{43.71} & 91.94 \\
Both (no fusion) & 0.251 & 0.729 & 46.33 & \underline{53.09} \\
\midrule
\ours\ (ours)    & \textbf{0.297} & \textbf{0.783} & \textbf{40.21} & \textbf{47.10} \\
\bottomrule
\end{tabular}
\end{table}

The four variants in Table \ref{tab:abl_fusion} expose a clear specialization–generalization tradeoff. Each single-channel ablation dominates the tasks aligned with its inductive bias: the geometric-only encoder has advantage on similarity (0.286 HR@1) and classification (0.773 macro-F1), both driven by \emph{where} a trajectory goes, while the kinematic-only encoder wins on prefix ETA (43.71\,s MAE), which depends on \emph{how} it moves. Removing either channel is catastrophic on the complementary regime—kinematic-only collapses on similarity to near-chance (0.028 HR@1), and both single-channel variants lose 40+\,s of MAE on full-trajectory TTE, a task that jointly requires spatial context (route) and motion dynamics (pace). The factorized-no-fusion variant shows that merely exposing the task head to both streams is insufficient: it underperforms each specialist on its specialty and still trails the full model on TTE by nearly 6\,s, indicating that isolated channels fail to exploit their complementarity. Cross-attention fusion is what closes this gap. Full \ours\ finishes the best on every specialist task and decisively wins on TTE (47.10\,s), the one task whose information requirement spans both regimes. This is the profile we aim for in a general-purpose trajectory encoder: near or surpass specialist quality on each single-regime task, and strict improvement wherever the two regimes must be combined.

\subsubsection{Pretraining Recipe}
\label{sec:abl_pretrain}

Finally, we examine how pretraining hyperparameters affect downstream transfer. Table~\ref{tab:abl_pretrain} tries three mask ratio at matched 60k steps and additionally reports a 120k continuation of our main configuration. 

\begin{table}[h]
\centering
\caption{Pretraining ablation. Mask ratio is swept at 60k steps; the last row extends the chosen recipe to 120k steps. Validation loss is computed on held-out pretraining data; retrieval metrics are zero-shot DTW similarity on the Porto 1k/10k bank. Best 60k result in \textbf{bold}. The 120k row shows that extending pretraining continues to lower validation loss but \emph{regresses} on every retrieval metric, consistent with our finding that pretraining loss alone is not a sufficient model-selection signal.}
\label{tab:abl_pretrain}
\small
\setlength{\tabcolsep}{3pt}
\begin{tabular}{lccccccc}
\toprule
Recipe & Mask ratio & Training Budget & Val loss $\downarrow$ & HR@1 $\uparrow$ & HR@10 $\uparrow$ & MRR $\uparrow$ & NDCG@10 $\uparrow$ \\
\midrule
Mild masking        & 0.25 & 60k  & 0.095 & 0.296 & 0.696 & 0.428 & 0.455 \\
Strong masking      & 0.35 & 60k  & 0.098 & 0.308 & 0.689 & 0.437 & 0.459 \\
More Budget     & 0.30 & 120k & \textbf{0.073} & 0.257 & 0.624 & 0.377 & 0.410 \\
\midrule
\textit{Ours} & 0.30 & 60k & 0.098 & \textbf{0.351} & \textbf{0.781} & \textbf{0.503} & \textbf{0.521} \\
\bottomrule
\end{tabular}
\end{table}

\paragraph{Mask ratio has a clear sweet spot.} Within the 60k comparison, mask ratio 0.30 yields the strongest downstream retrieval, outperforming both milder (0.25) and stronger (0.35) corruption. Too little masking produces a weak self-supervised signal; too much destroys the contextual structure needed for successful prediction. Our current setting balances these forces.

\paragraph{Pretraining loss is not a \textit{free} signal.} Extending the main recipe from 60k to 120k steps reduces validation loss by 25\% (0.098 $\to$ 0.073), yet zero-shot retrieval \emph{degrades} sharply from 0.351 to 0.257. 
This discrepancy points to a pretraining--transfer mismatch: lower masked-token loss does not necessarily imply a more transferable trajectory representation. Appendix~\ref{app:pretraining} provides additional analysis of this objective--transfer gap.

\section{Limitations}
\label{sec:lim}

All of our experiments are conducted on the Porto taxi dataset, which is a commonly used benchmark in trajectory representation learning and supports direct comparison with prior work. It would be interesting to continue exploring: i) transfer learning capability to other dataset with minimal or zero finetuning; ii) a larger scale of pretraining with multiple cities and trajectories with more variety. 
So far we have verified our learned trajectory representations on two different domains across four downstream applications; we believe it is possible to include more tasks in the future to extend implications beyond what we have covered. 
Finally, TrajTok relies solely on raw GPS signals; a promising direction is to enrich the spatial tokens with lightweight geographic priors (e.g., map and road-network features) directly at the token level, without reverting to costly map-matching preprocessing.

\section{Conclusion}
We presented TrajTok, a trajectory representation learning framework that converts raw, noisy GPS traces into density-adaptive multi-resolution spatial tokens and learns transferable representations through a factorized transformer encoder with geometric--kinematic specialization, cross-attention fusion, and simple masked pretraining. Across four downstream applications spanning both geometry-dominated and motion-sensitive settings, a single frozen pretrained encoder with lightweight task heads achieved strong transfer performance, outperforming or matching specialized prior methods on trajectory similarity search, call-type classification, prefix-based ETA prediction, and full-trajectory travel-time regression. Our ablations further showed that adaptive tokenization, factorized dual-channel modeling, and careful pretraining design each contribute materially to downstream generalization. Overall, these results suggest that learned multi-resolution spatial tokenization, combined with a simple yet effective masked objective, is a promising recipe for general-purpose trajectory foundation models, and motivate future work on scaling to broader geographies, longer trajectories, and more diverse mobility tasks.

% reference list
\bibliographystyle{plainnat}
\bibliography{base}

@inproceedings{najjar2023towards,
  title={Towards a foundation model for trajectory intelligence},
  author={Najjar, Alameen},
  booktitle={2023 IEEE International Conference on Data Mining Workshops (ICDMW)},
  pages={832--835},
  year={2023},
  organization={IEEE}
}

@inproceedings{jiang2023self,
  title={Self-supervised trajectory representation learning with temporal regularities and travel semantics},
  author={Jiang, Jiawei and Pan, Dayan and Ren, Houxing and Jiang, Xiaohan and Li, Chao and Wang, Jingyuan},
  booktitle={2023 IEEE 39th international conference on data engineering (ICDE)},
  pages={843--855},
  year={2023},
  organization={IEEE}
}

@inproceedings{unitraj2025,
  title={UniTraj: Learning a Universal Trajectory Foundation Model from Billion-Scale Worldwide Traces},
  author={Zhu, Yuanshao and Yu, James Jianqiao and Zhao, Xiangyu and Zhou, Xun and Han, Liang and Wei, Xuetao and Liang, Yuxuan},
  booktitle={The Thirty-ninth Annual Conference on Neural Information Processing Systems},
  year={2025}
}

@article{zheng2014urban,
  title={Urban computing: concepts, methodologies, and applications},
  author={Zheng, Yu and Capra, Licia and Wolfson, Ouri and Yang, Hai},
  journal={ACM Transactions on Intelligent Systems and Technology (TIST)},
  volume={5},
  number={3},
  pages={1--55},
  year={2014},
  publisher={ACM New York, NY, USA}
}

@article{su2024roformer,
  title={Roformer: Enhanced transformer with rotary position embedding},
  author={Su, Jianlin and Ahmed, Murtadha and Lu, Yu and Pan, Shengfeng and Bo, Wen and Liu, Yunfeng},
  journal={Neurocomputing},
  volume={568},
  pages={127063},
  year={2024},
  publisher={Elsevier}
}

@inproceedings{space2vec,
  title={Multi-Scale Representation Learning for Spatial Feature Distributions using Grid Cells},
  author={Mai, Gengchen and Janowicz, Krzysztof and Yan, Bo and Zhu, Rui and Cai, Ling and Lao, Ni},
  booktitle={8th International Conference on Learning Representations, ICLR 2020},
  year={2020}
}

@inproceedings{didi_dispatching,
  title={Large-scale order dispatch in on-demand ride-hailing platforms: A learning and planning approach},
  author={Xu, Zhe and Li, Zhixin and Guan, Qingwen and Zhang, Dingshui and Li, Qiang and Nan, Junxiao and Liu, Chunyang and Bian, Wei and Ye, Jieping},
  booktitle={Proceedings of the 24th ACM SIGKDD international conference on knowledge discovery \& data mining},
  pages={905--913},
  year={2018}
}

@article{nature_mobility_covid,
  title={Mobility network models of COVID-19 explain inequities and inform reopening},
  author={Chang, Serina and Pierson, Emma and Koh, Pang Wei and Gerardin, Jaline and Redbird, Beth and Grusky, David and Leskovec, Jure},
  journal={Nature},
  volume={589},
  number={7840},
  pages={82--87},
  year={2021},
  publisher={Nature Publishing Group UK London}
}

@inproceedings{li2018deep,
  title     = {Deep Representation Learning for Trajectory Similarity Computation},
  author    = {Li, Xiucheng and Zhao, Kaiqi and Cong, Gao and Jensen, Christian S. and Wei, Wei},
  booktitle = {Proceedings of the 34th IEEE International Conference on Data Engineering (ICDE)},
  pages     = {617--628},
  year      = {2018},
  publisher = {IEEE}
}

@inproceedings{deng2022efficient,
  title     = {Efficient Trajectory Similarity Computation with Contrastive Learning},
  author    = {Deng, Liwei and Zhao, Yan and Wang, Zidan and Fan, Hao and Zheng, Kai},
  booktitle = {Proceedings of the 31st ACM International Conference on Information \& Knowledge Management (CIKM)},
  pages     = {229--239},
  year      = {2022},
  publisher = {ACM},
  doi       = {10.1145/3511808.3557308}
}

@inproceedings{yao2019computing,
  title     = {Computing Trajectory Similarity in Linear Time: A Generic Seed-Guided Neural Metric Learning Approach},
  author    = {Yao, Di and Zhang, Chao and Zhu, Zhihua and Hu, Jianhui and Bi, Jingping},
  booktitle = {Proceedings of the 35th IEEE International Conference on Data Engineering (ICDE)},
  pages     = {1358--1369},
  year      = {2019},
  publisher = {IEEE},
  doi       = {10.1109/ICDE.2019.00123}
}

@inproceedings{chang2023contrastive,
  title     = {Contrastive Trajectory Similarity Learning with Dual-Feature Attention},
  author    = {Chang, Yanchuan and Qi, Jianzhong and Liang, Yuxuan and Tanin, Egemen},
  booktitle = {Proceedings of the 39th IEEE International Conference on Data Engineering (ICDE)},
  year      = {2023},
  publisher = {IEEE},
  doi       = {10.1109/ICDE55515.2023.00119}
}

@inproceedings{yao2022trajgat,
  title     = {{TrajGAT}: A Graph-based Long-term Dependency Modeling Approach for Trajectory Similarity Computation},
  author    = {Yao, Di and Hu, Haonan and Du, Lun and Cong, Gao and Han, Shi and Bi, Jingping},
  booktitle = {Proceedings of the 28th ACM SIGKDD Conference on Knowledge Discovery and Data Mining (KDD)},
  pages     = {2275--2285},
  year      = {2022},
  publisher = {ACM},
  doi       = {10.1145/3534678.3539358}
}

@inproceedings{chen2024kgts,
  title     = {{KGTS}: Contrastive Trajectory Similarity Learning over Prompt Knowledge Graph Embedding},
  author    = {Chen, Zhen and Zhang, Dalin and Feng, Shanshan and Chen, Kaixuan and Chen, Lisi and Han, Peng and Shang, Shuo},
  booktitle = {Proceedings of the 38th AAAI Conference on Artificial Intelligence (AAAI)},
  year      = {2024},
  publisher = {AAAI Press},
  doi       = {10.1609/aaai.v38i8.28672}
}

@article{yang2025simformer,
  title   = {{SIMformer}: Single-Layer Vanilla Transformer Can Learn Free-Space Trajectory Similarity},
  author  = {Yang, Chuang and Jiang, Renhe and Xu, Xiaohang and Xiao, Chuan and Sezaki, Kaoru},
  journal = {Proceedings of the VLDB Endowment},
  year    = {2025},
  doi     = {10.14778/3705829.3705853}
}

@article{schestakov2024trajectory,
  title={Trajectory representation learning on road networks and grids with spatio-temporal dynamics},
  author={Schestakov, Stefan and Gottschalk, Simon},
  journal={arXiv preprint arXiv:2411.14014},
  year={2024}
}

@article{lin2023pretraining,
  title   = {Pre-training General Trajectory Embeddings with Maximum Multi-view Entropy Coding},
  author  = {Lin, Yan and Wan, Huaiyu and Guo, Shengnan and Hu, Jilin and Jensen, Christian S. and Lin, Youfang},
  journal = {IEEE Transactions on Knowledge and Data Engineering},
  year    = {2023},
  doi     = {10.48550/arXiv.2207.14539}
}

@article{yao2018learning,
author = {Yao, Di and Zhang, Chao and Zhu, Zhihua and Hu, Qin and Wang, Zheng and Huang, Jianhui and Bi, Jingping},
title = {Learning deep representation for trajectory clustering},
journal = {Expert Systems},
volume = {35},
number = {2},
pages = {e12252},
doi = {https://doi.org/10.1111/exsy.12252},
url = {https://onlinelibrary.wiley.com/doi/abs/10.1111/exsy.12252},
eprint = {https://onlinelibrary.wiley.com/doi/pdf/10.1111/exsy.12252},
note = {e12252 10.1111/exsy.12252},
year = {2018}
}

@article{yang2021unsupervised,
  title={Unsupervised path representation learning with curriculum negative sampling},
  author={Yang, Sean Bin and Guo, Chenjuan and Hu, Jilin and Tang, Jian and Yang, Bin},
  journal={arXiv preprint arXiv:2106.09373},
  year={2021}
}

@inproceedings{ma2024more,
  title={More than routing: Joint GPS and route modeling for refine trajectory representation learning},
  author={Ma, Zhipeng and Tu, Zheyan and Chen, Xinhai and Zhang, Yan and Xia, Deguo and Zhou, Guyue and Chen, Yilun and Zheng, Yu and Gong, Jiangtao},
  booktitle={Proceedings of the ACM Web Conference 2024},
  pages={3064--3075},
  year={2024}
}

@misc{yang2024fineline,
  author = {Chen Yang and Junzhuo Li and Xinyao Niu and Xinrun Du and Songyang Gao and Haoran Zhang and Zhaoliang Chen and Xingwei Qu and Ruibin Yuan and Yizhi Li and Jiaheng Liu and Stephen W. Huang and Shawn Yue and Ge Zhang},
  title = {The Fine Line: Navigating Large Language Model Pretraining with Down-streaming Capability Analysis},
  year = {2024},
  eprint = {2404.01204},
  archivePrefix = {arXiv},
  primaryClass = {cs.CL},
  url = {https://arxiv.org/abs/2404.01204},
  note = {Direct support: the abstract explicitly states that pretraining-loss-based scaling focuses on compression on training data and can be inconsistent with improvements on downstream tasks.}
}

@misc{chen2026nexus,
  author = {Huanran Chen and Huaqing Zhang and Xiao Li and Yinpeng Dong and Ke Shen and Jun Zhu},
  title = {Nexus: Same Pretraining Loss, Better Downstream Generalization via Common Minima},
  year = {2026},
  eprint = {2604.09258},
  archivePrefix = {arXiv},
  primaryClass = {cs.LG},
  url = {https://arxiv.org/abs/2604.09258},
  note = {Direct support: the paper reports significantly better downstream performance despite achieving the same pretraining loss, challenging reliance on pretraining loss as the sole evaluation proxy.}
}

@misc{uberh3,
  author = {{Uber Technologies}},
  title = {{H3: A Hexagonal Hierarchical Geospatial Indexing System (version 4)}},
  year = {2022},
  howpublished = {\url{https://h3geo.org/}},
  note = {Accessed: April 22, 2026}
}

@article{vaswani2017attention,
  title={Attention is all you need},
  author={Vaswani, Ashish and Shazeer, Noam and Parmar, Niki and Uszkoreit, Jakob and Jones, Llion and Gomez, Aidan N and Kaiser, {\L}ukasz and Polosukhin, Illia},
  journal={Advances in neural information processing systems},
  volume={30},
  year={2017}
}

@inproceedings{wang2018deepTTE,
  title     = {When Will You Arrive? Estimating Travel Time Based on Deep Neural Networks},
  author    = {Wang, Dong and Zhang, Junbo and Cao, Wei and Li, Jian and Zheng, Yu},
  booktitle = {Proceedings of the Thirty-Second AAAI Conference on Artificial Intelligence},
  pages     = {2500--2507},
  year      = {2018}
}

@inproceedings{fu2020compacteta,
  title     = {{CompactETA}: A Fast Inference System for Travel Time Prediction},
  author    = {Fu, Kun and Meng, Fanlin and Ye, Jieping and Wang, Zheng},
  booktitle = {Proceedings of the 26th ACM SIGKDD International Conference on Knowledge Discovery and Data Mining},
  pages     = {3337--3345},
  year      = {2020},
  doi       = {10.1145/3394486.3403387}
}

@inproceedings{fang2020constgat,
  title     = {{ConSTGAT}: Contextual Spatial-Temporal Graph Attention Network for Travel Time Estimation at Baidu Maps},
  author    = {Fang, Xiaomin and Huang, Jizhou and Wang, Fan and Zeng, Lingke and Liang, Haijin and Wang, Haifeng},
  booktitle = {Proceedings of the 26th ACM SIGKDD International Conference on Knowledge Discovery and Data Mining},
  pages     = {2697--2705},
  year      = {2020},
  doi       = {10.1145/3394486.3403316}
}

@inproceedings{fan2022metaer,
  title     = {{MetaER-TTE}: An Adaptive Meta-Learning Model for En Route Travel Time Estimation},
  author    = {Fan, Yanyan and Yuan, Ye and Lv, Guoliang and Chen, Hao and Li, Jian and Liu, Feifei},
  booktitle = {Proceedings of the Thirty-First International Joint Conference on Artificial Intelligence},
  pages     = {1806--1812},
  year      = {2022},
  doi       = {10.24963/ijcai.2022/251}
}

@inproceedings{finn2017maml,
  title     = {Model-Agnostic Meta-Learning for Fast Adaptation of Deep Networks},
  author    = {Finn, Chelsea and Abbeel, Pieter and Levine, Sergey},
  booktitle = {Proceedings of the 34th International Conference on Machine Learning},
  pages     = {1126--1135},
  year      = {2017}
}

@inproceedings{fang2021ssml,
  title     = {{SSML}: Self-Supervised Meta-Learner for En Route Travel Time Estimation at Baidu Maps},
  author    = {Fang, Xiaomin and Huang, Jizhou and Wang, Fan and Liu, Lihang and Sun, Yibo and Wang, Haifeng},
  booktitle = {Proceedings of the 27th ACM SIGKDD Conference on Knowledge Discovery and Data Mining},
  pages     = {2840--2848},
  year      = {2021},
  doi       = {10.1145/3447548.3467109}
}

@article{long2026region, title={Region-Point Joint Representation for Effective Trajectory Similarity Learning}, volume={40}, url={https://ojs.aaai.org/index.php/AAAI/article/view/38571}, DOI={10.1609/aaai.v40i18.38571}, abstractNote={Recent learning-based methods have reduced the computational complexity of traditional trajectory similarity computation, but state-of-the-art (SOTA) methods still fail to leverage the comprehensive spectrum of trajectory information for similarity modeling. To tackle this problem, we propose RePo, a novel method that jointly encodes Region-wise and Point-wise features to capture both spatial context and fine-grained moving patterns. For region-wise representation, the GPS trajectories are first mapped to grid sequences, and spatial context are captured by structural features and semantic context enriched by visual features. For point-wise representation, three lightweight expert networks extract local, correlation, and continuous movement patterns from dense GPS sequences. Then, a router network adaptively fuses the learned point-wise features, which are subsequently combined with region-wise features using cross-attention to produce the final trajectory embedding. To train RePo, we adopt a contrastive loss with hard negative samples to provide similarity ranking supervision. Experiment results show that RePo achieves an average accuracy improvement of 22.2% over SOTA baselines across all evaluation metrics.}, number={18}, journal={Proceedings of the AAAI Conference on Artificial Intelligence}, author={Long, Hao and Zhou, Silin and Chen, Lisi and Shang, Shuo}, year={2026}, month={Mar.}, pages={15439-15447} }

@article{mai2020multi,
  title={Multi-scale representation learning for spatial feature distributions using grid cells},
  author={Mai, Gengchen and Janowicz, Krzysztof and Yan, Bo and Zhu, Rui and Cai, Ling and Lao, Ni},
  journal={arXiv preprint arXiv:2003.00824},
  year={2020}
}

@inproceedings{klemmer2025satclip,
  title={Satclip: Global, general-purpose location embeddings with satellite imagery},
  author={Klemmer, Konstantin and Rolf, Esther and Robinson, Caleb and Mackey, Lester and Ru{\ss}wurm, Marc},
  booktitle={Proceedings of the AAAI Conference on Artificial Intelligence},
  volume={39},
  number={4},
  pages={4347--4355},
  year={2025}
}

\appendix

\section{Algorithm: Density-Adaptive Vocabulary Construction}

\begin{algorithm}[H]
  \caption{Density-Adaptive Vocabulary Construction}
  \label{alg:adaptive_vocab}
  \begin{algorithmic}[1]
  \Require
    point stream $\mathcal{P}$;
    base resolution $r_{\min}$;
    max resolution $r_{\max}$;
    capacity threshold $C$
  \Ensure
    vocabulary $\mathcal{V}$ of (cell\_id, resolution) pairs

  \ForAll{$p \in \mathcal{P}$}
    \State $c \gets \textsc{H3Cell}(\textsc{LatLon}(p),\, r_{\min})$
    \State $\texttt{count}[c] \mathrel{+}= 1$
  \EndFor

  \State $Q \gets \{(c,\, r_{\min},\, \texttt{count}[c]) : \texttt{count}[c] > C\}$
  \State $\mathcal{V} \gets \{c : \texttt{count}[c] \leq C\}$

  \While{$Q \neq \emptyset$}
    \State $(c,\, r,\, S) \gets \textsc{Pop}(Q)$
    \If{$|S| > C$ \textbf{and} $r < r_{\max}$}
      \ForAll{$c' \in \textsc{Children}(c,\, r+1)$}
        \State $S' \gets \{p \in S : \textsc{H3Cell}(p,\, r+1) = c'\}$
        \State push $(c',\, r+1,\, S')$ into $Q$
      \EndFor
    \Else
      \State $\mathcal{V} \gets \mathcal{V} \cup \{c\}$
    \EndIf
  \EndWhile

  \State \Return $\mathcal{V}$
  \end{algorithmic}
\end{algorithm}

\section{Implementation Details}
\label{app:impl}

\subsection{Experiment Setup}
\label{app:setup}

\paragraph{Dataset.} We use the Porto taxi dataset (1{,}710{,}670 trips). After dropping incomplete data points and points outside the reasonable bounding box $[41.100, 41.220] \times [-8.700, -8.530]$, a deterministic 60/20/20 hash-based split yields 885{,}055 / 294{,}909 / 294{,}367 train/validation/test trajectories. 
% Full preprocessing and split-restoration details are in Appendix~\ref{app:data}.

\paragraph{Spacial Vocabulary.} Our adaptive vocabulary uses base H3 resolution 6, maximum resolution 9, and capacity 1{,}000 data points, yielding 1{,}494 cells. We visualize the vocabulary in Figure~\ref{fig:tokenization}. 

\paragraph{Model.} The factorized encoder has 144M parameters: $d_{\text{model}}{=}512$, 8 heads, 16 total layers with 4 cross-attention fusion layers, $\texttt{max\_seq\_len}{=}192$, and operates on raw token streams. Since each attention head has dimension 64, we allocate $(20,20,24)$ rotated dimensions to latitude, longitude, and time. Relative latitude and longitude are multiplied by $10^4$ before forming rotary angles.

\paragraph{Hyperparameters.} Factorized masked objective (§\ref{sec:pretraining}) with mask ratio 0.3, run-aware span masking ($\ell{=}6$), AdamW ($\text{lr}{=}10^{-4}$, wd${=}0.01$), cosine decay with 20\% warmup, bf16, effective batch size 256, for 60k steps. Motion loss weights $\beta_{\text{speed}}{=}\beta_{\text{heading}}{=}1.0$. For hyperparameter ablations, please see §\ref{sec:ablations}.

\paragraph{Computing Resources.} All experiments ran on a single GPU (NVIDIA RTX 6000 Ada Generation cards with 48 GiB VRAM). Pretraining the main TrajTok dual-stream model takes roughly 1.5-2 hours per run; downstream finetuning for similarity, classification, ETA, and TTE completes in 30–60 minutes per run with the encoder frozen.

\subsection{Downstream Adapter Configurations}
\label{app:adapters}

Throughout §\ref{sec:sim}--§\ref{sec:tte} the pretrained \ours\ encoder
is frozen and only a lightweight, task-specific adapter is trained on
top. Below we detail the adapter architecture, training data, and
optimization settings used for each benchmark.

\paragraph{Trajectory Similarity Search (§\ref{sec:sim}).}
The retrieval head is an attentive pooling layer followed by a linear
projection, applied to the frozen encoder output and L2-normalized for
cosine-similarity ranking. It is trained on a separate train bank of
2{,}048 queries against 8{,}192 corpus trajectories, disjoint from the
evaluation bank (1{,}000 queries against a 10{,}000-trajectory corpus).
The objective combines InfoNCE contrastive loss with a rank-distillation
term against DTW ground truth, with early stopping on validation HR@1.

\paragraph{Trajectory Classification (§\ref{sec:cls}).}
The call-type classifier concatenates an attentive-pooled trajectory
embedding with a compact departure-context branch consisting of the
start token, start location, and a 32-dim origin-time interaction
feature. Training uses AdamW ($\text{lr}{=}3{\times}10^{-3}$,
$\text{wd}{=}0.05$) with batch size 4{,}096.

\paragraph{ETA from Prefixes (§\ref{sec:eta}).}
An attentive pooling layer conditioned on the destination token
aggregates the prefix-trajectory representation, which is then mapped to
the remaining travel time by a lightweight regressor. Training uses a
randomized-prefix regime in which the observed prefix ratio is sampled
per example as $|\text{prefix}| \sim U[40\%, 60\%]$ so that the head is
exposed to a wide range of completion ratios.

\paragraph{Full-Trajectory Travel-Time Regression (§\ref{sec:tte}).}
The full-trip regressor mirrors the ETA adapter (attentive pooling
followed by a lightweight regressor) but consumes the entire observed
trajectory and omits prefix sampling.

\subsection{Attention}
\label{app:attention}

This subsection gives formal definitions of the self-attention and cross-attention
operations referenced in \S3.4 and equations~(3)--(6) of the main text. Both are
instantiated as multi-head scaled dot-product attention~\citep{vaswani2017attention}
with rotary position embeddings (RoPE)~\citep{su2024roformer} applied to the per-head
query and key projections.

\paragraph{Scaled dot-product attention.}
Given queries $Q \in \mathbb{R}^{L_q \times d_h}$, keys
$K \in \mathbb{R}^{L_k \times d_h}$, and values $V \in \mathbb{R}^{L_k \times d_h}$,
scaled dot-product attention is defined as
\begin{equation}
\mathrm{Attn}(Q, K, V) \;=\; \mathrm{softmax}\!\left(\frac{Q K^{\top}}{\sqrt{d_h}}\right) V.
\end{equation}
Multi-head attention concatenates $H$ parallel heads of dimension $d_h = d/H$:
\begin{equation}
\mathrm{MHA}\!\left(X_Q, X_K, X_V;\, \mathbf{P}_Q, \mathbf{P}_K\right)
\;=\; \mathrm{Concat}\!\left(\mathrm{head}_1, \ldots, \mathrm{head}_H\right) W^{O},
\end{equation}
with
\begin{equation}
\mathrm{head}_h \;=\; \mathrm{Attn}\!\left(
    \mathcal{R}\!\left(X_Q W_h^{Q};\, \mathbf{P}_Q\right),\;
    \mathcal{R}\!\left(X_K W_h^{K};\, \mathbf{P}_K\right),\;
    X_V W_h^{V}\right),
\end{equation}
where $W_h^{Q}, W_h^{K}, W_h^{V} \in \mathbb{R}^{d \times d_h}$,
$W^{O} \in \mathbb{R}^{d \times d}$, and $\mathcal{R}(\cdot;\, \mathbf{p})$ denotes
the RoPE rotation conditioned on positional coordinates $\mathbf{p}$.

\paragraph{Rotary position embedding.}
For a token $j$ in the \emph{kinematic} channel, $\mathbf{p}_j = t_j$ is the
trajectory-relative timestamp, and $\mathcal{R}$ applies a one-dimensional temporal
rotation across all $d_h$ head dimensions. For a token in the \emph{geometric}
channel, $\mathbf{p}_j = (\phi_j, \lambda_j, t_j)$ is a three-axis spatiotemporal
tuple; the head is partitioned into disjoint blocks of sizes
$(d_\phi, d_\lambda, d_t)$ with $d_\phi + d_\lambda + d_t = d_h$, and each
coordinate rotates its corresponding block independently. In our implementation,
$d_h = 64$ and $(d_\phi, d_\lambda, d_t) = (20, 20, 24)$; relative latitudes and
longitudes are multiplied by $10^{4}$ before forming rotary angles.

\paragraph{Self-attention.}
Let $X^{(\ell)} \in \mathbb{R}^{L \times d}$ denote the hidden states of a single
channel at layer $\ell$, associated with per-token positional coordinates
$\mathbf{P} = \{\mathbf{p}_j\}_{j=1}^{L}$. Self-attention draws queries, keys, and
values from the same input:
\begin{equation}
\mathrm{SelfAttn}\!\left(X^{(\ell)};\, \mathbf{P}\right) \;=\;
\mathrm{MHA}\!\left(X^{(\ell)},\, X^{(\ell)},\, X^{(\ell)};\; \mathbf{P}, \mathbf{P}\right).
\end{equation}
The two streams differ only in the choice of $\mathbf{P}$:
$\mathrm{SelfAttn}_{\mathrm{geo}}$ uses the spatiotemporal tuple
$\mathbf{p}_j = (\phi_j, \lambda_j, t_j)$, while $\mathrm{SelfAttn}_{\mathrm{kin}}$
uses the temporal coordinate $\mathbf{p}_j = t_j$ alone.

\paragraph{Cross-attention.}
In a fusion block, each stream queries the other. Given two channels
$X^{(\ell)}, Y^{(\ell)} \in \mathbb{R}^{L \times d}$ over the same token positions,
cross-attention on $X$ with context $Y$ is
\begin{equation}
\mathrm{CrossAttn}\!\left(X^{(\ell)}, Y^{(\ell)};\, \mathbf{P}\right) \;=\;
\mathrm{MHA}\!\left(X^{(\ell)},\, Y^{(\ell)},\, Y^{(\ell)};\; \mathbf{P}, \mathbf{P}\right),
\end{equation}
i.e.\ queries are projected from $X$ while keys and values are projected from $Y$.
Because fusion layers align geometric and kinematic tokens in a shared spatiotemporal
frame, both $\mathrm{CrossAttn}_{\mathrm{geo}}$ and $\mathrm{CrossAttn}_{\mathrm{kin}}$
use the spatiotemporal coordinates $\mathbf{p}_j = (\phi_j, \lambda_j, t_j)$.

Let $G^{(\ell)}$ and $K^{(\ell)}$ denote the geometric and kinematic hidden states at layer $\ell$. We have:

\begin{align}
G^{(\ell+1)} &\;=\; \mathrm{CrossAttn}\!\left(\tilde{G}^{(\ell)},\, K^{(\ell)};\, \mathbf{P}\right),\\
K^{(\ell+1)} &\;=\; \mathrm{CrossAttn}\!\left(\tilde{K}^{(\ell)},\, G^{(\ell)};\, \mathbf{P}\right).
\end{align}

\paragraph{Block structure.}
Every attention sublayer is wrapped in a pre-norm residual block followed by a
two-layer GeGLU MLP, consistent with Figure~\ref{fig:overview} (Middle):
\begin{align}
X' &\;=\; X + \mathrm{SelfAttn}\!\left(\mathrm{LN}(X);\, \mathbf{P}\right),\\
X'' &\;=\; X' + \mathrm{MLP}\!\left(\mathrm{LN}(X')\right),
\end{align}
with analogous pre-norm wrapping for the cross-attention sublayers in the fusion blocks.

\subsection{ETA Baselines}
\label{app:eta_base_feats}

For \textit{Hoch-style gradient-boosted regressor} (LightGBM, Huber loss, \texttt{log1p} target, up to 3000 trees with early stopping on validation MAE), we manually derieved the following features: calendar context (weekday, hour, sinusoidal time-of-day), prefix shape (length, duration, cumulative and straight-line distance), current state (latitude, longitude, speed, heading, distance-to-city-center), recent-window means over the last five points, and destination conditioning (destination coordinates, Haversine from current point to destination, bearing to destination). This follows the spirit of the ECML/PKDD 2015 Porto taxi-ETA challenge winner, extended with the destination cues that match our protocol. 

For \textit{prefix-cell × destination-cell kNN}, we index every training prefix by the pair of H3 cells at resolution 9 covering its prefix-last point and its destination, take a $k{=}10$ mean of matching remaining-time targets, and fall back through H3-r8, H3-r7, destination-cell-only, and global-mean when the exact bucket is sparse. On our test set, 95.4\% of queries resolve at exact H3-r9, indicating that the memorization baseline is genuinely exercised and not dominated by fallbacks.

\section{Supplementary Ablations}
\label{app:abl}

This appendix reports additional ablation studies that are included for completeness. These studies cover (i) masking-strategy variants, (ii) a deduplication control, and (iii) a classification-head design study. Unless otherwise noted, all experiments use the same dual-stream encoder, pretraining objective, and evaluation protocols as the main paper.

\subsection{Masking Strategy: Naive vs. Run-Aware Span Masking}
\label{app:abl-mask}

Raw token streams often contain repeated-cell runs, which can make naive span masking degenerate. In particular, when a masked span covers only the interior of a constant-cell block, the missing tokens are trivially recoverable from the visible endpoints. Our main recipe therefore uses run-aware span masking, which keeps the masked objectives unchanged but avoids this shortcut in the corruption process.

For a length-$L$ sequence, we set the mask budget to $\lfloor \rho L \rfloor$ and sample candidate span lengths uniformly from $\{\ell-1,\ell,\ell+1\}$ around the average span length $\ell$. A candidate span is rejected if both endpoints lie strictly inside the same repeated-cell run, meaning that the span would mask only the interior of one constant-cell block. After rejection sampling, any remaining budget is filled first from non-interior tokens and only then, if necessary, from arbitrary valid tokens. Table~\ref{tab:app_masking} compares this run-aware strategy with naive span masking under matched 12k-step budgets.

\begin{table}[h]
\centering
\caption{Masking strategy comparison at matched 12k training steps. Both rows use raw token streams, mask ratio 0.30, and identical encoder configurations.}
\label{tab:app_masking}
\small
\begin{tabular}{lcccc}
\toprule
Masking strategy & Val loss $\downarrow$ & HR@1 $\uparrow$ & MRR $\uparrow$ & Spearman $\uparrow$ \\
\midrule
Naive              & \textbf{0.276} & 0.286 & 0.424 & 0.524 \\
Run-aware (ours)   & 0.297 & \textbf{0.308} & \textbf{0.442} & \textbf{0.548} \\
\bottomrule
\end{tabular}
\end{table}

The two strategies show clear distinctions on pretraining loss score and downstream performance: while naive span masking does provide a lower loss in general, the representations learned are outperformed by our run-aware strategy, which indicates that run-aware masking prevent the model from taking the shortcuts by simply copying surrounding tokens and incentivize it to actually learn from high level geometric features.

\subsection{Deduplicated vs. Raw Token Sequences}
\label{app:abl-dedup}

Our main backbone pretrains on complete token sequence, which preserve dwell patterns and repeated-cell motion cues at the cost of longer sequences. We also tried another de-duplicate design which collapse consecutive repeated cells. Table~\ref{tab:app_dedup} reports the deduplicated control for transparency.

\begin{table}[h]
\centering
\caption{Deduplicated vs. raw token streams. The two configurations differ in both sequence length and masking strategy, so this comparison is not fully matched.}
\label{tab:app_dedup}
\small
\begin{tabular}{lcccc}
\toprule
Configuration &  Val loss $\downarrow$ & HR@1 $\uparrow$ & MRR $\uparrow$ \\
\midrule
Dedup          & 0.098 & 0.309 & 0.459 \\
Ours   & \textbf{0.047} & \textbf{0.351} & \textbf{0.588} \\
\bottomrule
\end{tabular}
\end{table}

\subsection{Call-Type Classification Head Design}
\label{app:abl-cls}

The main paper reports call-type classification results using our best head configuration: attentive-pooled encoder features concatenated with a compact departure-context branch (start token, start location, and a 32-dim origin-time interaction). Here we report variants of this head, holding the frozen encoder fixed. This ablation studies the \emph{task head} rather than the encoder, and is included to document design decisions in the downstream evaluation protocol.

\begin{table}[h]
\centering
\caption{Call-type classification head ablation. All rows use the same frozen \ours\ encoder.}
\label{tab:app_cls_head}
\small
\begin{tabular}{lcc}
\toprule
Head variant & Test accuracy $\uparrow$ & Test macro-F1 $\uparrow$ \\
\midrule
Departure context + large origin-time branch & 0.740 & 0.692 \\
Departure context only                       & 0.804 & 0.767 \\
Departure context + compact origin-time branch (main) & \textbf{0.811} & \textbf{0.773} \\
\bottomrule
\end{tabular}
\end{table}

A large origin-time interaction branch over-parameterizes the head and degrades performance, while omitting origin-time context entirely sacrifices useful temporal signal. The compact 32-dim branch used in our main configuration balances these extremes. These differences reflect task-head design choices and do not affect the frozen encoder representation itself.

\subsection{Pretraining Recipe}
\label{app:pretraining}

During pretraining, although the training loss and validation loss both decrease steadily, the downstream performance (we use zero-shot similar trajectory search without finetuning as an indicator for simplicity), on each checkpoint, is not decreasing monotonically (see Figure~\ref{fig:pretraining_loss_curve}). We believe this pretraining-benchmark mismatch signals a deeper alignment research problem which is beyond our scope and require further study~\cite{yang2024fineline, chen2026nexus}.

\begin{figure}
    \centering
    \includegraphics[width=1.0\linewidth]{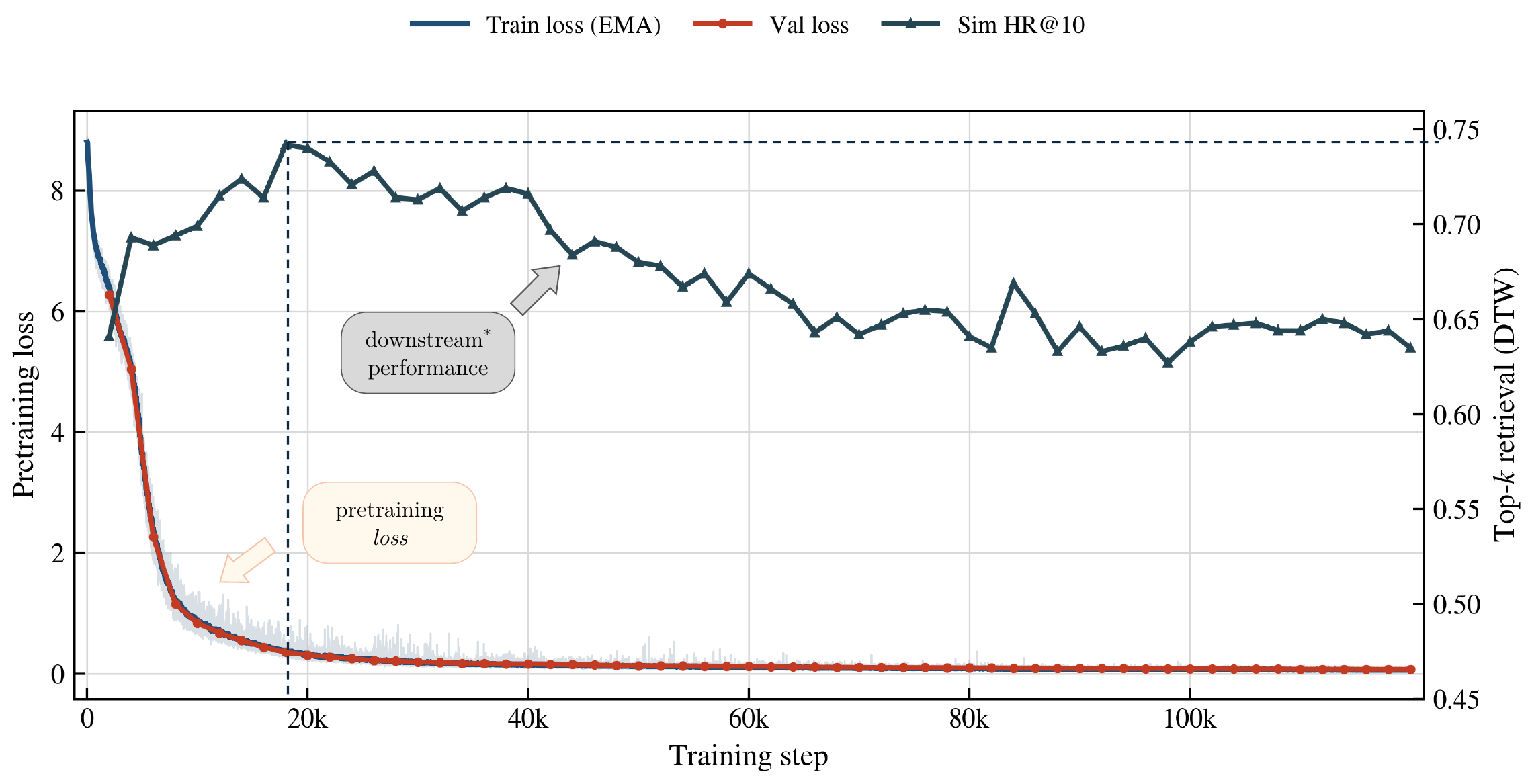}
    \caption{Mismatch between pretraining optimization and downstream utility. Although both training and validation pretraining losses decrease monotonically, downstream retrieval performance (Sim HR@10) peaks early and then gradually declines, indicating that lower pretraining loss does not necessarily yield better transfer performance.}
    \label{fig:pretraining_loss_curve}
\end{figure}

\section{Broader Impacts}
\label{app:social}

TrajTok is foundational research on representation learning for GPS trajectories, with positive potential applications in urban planning, public transit optimization, traffic management, ride-hailing efficiency, and epidemic modeling, all of which can improve mobility services and public health outcomes. However, we acknowledge that any technology that produces transferable representations of human movement carries privacy and surveillance risks: trajectory embeddings can facilitate re-identification of individuals from sparse location traces, enable behavioral profiling, and lower the cost of large-scale mobility tracking if applied to non-consented data. These risks apply whether the model is functioning correctly (intended downstream tasks may still aggregate sensitive mobility patterns), produces incorrect outputs (misclassification could propagate into downstream decisions affecting individuals or groups, e.g., in dispatch or pricing systems), or is misused (adversaries repurposing pretrained encoders for tracking or stalking). All experiments in this paper use the publicly released Porto taxi dataset, which contains anonymized vehicle (not individual user) trajectories collected and released for research, and we do not collect new mobility data. To mitigate misuse, we recommend that practitioners building on TrajTok apply standard privacy safeguards — including k-anonymity or differential privacy on training data, restricting access to pretrained checkpoints trained on sensitive corpora, and avoiding fine-grained re-identification tasks — and we will document these considerations in any code or model release.

\end{document}